\documentclass{article} 
\usepackage{iclr2020_conference,times}
\usepackage{mathtools}
\usepackage{amssymb}
\usepackage{amsfonts}
\everymath{\displaystyle}
\pdfoutput=1
\usepackage{mathtools}
\usepackage{amsmath,amsfonts,bm}

\usepackage{mathtools}
\usepackage{amsmath,amsfonts,bm}

\def\eqref#1{equation~\ref{#1}}

\def\1{\bm{1}}

\DeclareMathAlphabet{\mathsfit}{\encodingdefault}{\sfdefault}{m}{sl}
\SetMathAlphabet{\mathsfit}{bold}{\encodingdefault}{\sfdefault}{bx}{n}

\usepackage{hyperref}
\usepackage{xstring}
\newcommand{\method}{AMUSED}
\newcommand{\lstm}{LSTM}
\newcommand{\gru}{GRU}
\newcommand{\reffig}{Figure \ref}
\newcommand{\refsec}{Section \ref}
\newcommand{\refeqn}{Equation \ref}
\newcommand{\reftable}{Table \ref}
\newcommand{\norm}[1]{\left\lVert#1\right\rVert}

\usepackage{graphicx}
\usepackage{hyperref}
\usepackage[normalem]{ulem}
\usepackage{url}
\usepackage{xr}
\usepackage{booktabs}
\usepackage{adjustbox}
\newcommand*{\crosssymbol}{%
    \text{%
      \raise 1ex\hbox{%
        \rlap{\vrule height.2pt depth.2pt width .75ex}%
        \hbox to .75ex{\hss\vrule height .5ex depth 1ex\hss}%
      }%
    }%
}

\title{AMUSED: A Multi-Stream Vector Representation Method for Use in Natural Dialogue}

\author{Gaurav Kumar**, Promod Yenigalla \\
Voice Intelligence R\& D\\
Samsung Research Institute\\
Bangalore, India\\
\texttt{\{gaurav.k1, promod.y\}@samsung.com} \\
\And
Rishabh Joshi**\\
Language Technologies Institute\\
Carnegie Mellon University\\
Pittsburgh, PA, USA \\
\texttt{rjoshi2@andrew.cmu.edu} \\
\AND
Jaspreet Singh** \\
Department of Computer Science \\
Stony Brook University \\
Stony Brook, NY, USA\\
\texttt{jaspreet.ahluwalia@stonybrook.edu}
\thanks{Here, ** denotes equal contribution. The work was conducted at Samsung Research Institute, Bangalore, India} 
}

\iclrfinalcopy 
\begin{document}

\maketitle

\begin{abstract}
The problem of building a coherent and non-monotonous conversational agent with proper
discourse and coverage is still an area of open
research. Current architectures only take care
of semantic and contextual information for a
given query and fail to completely account for
syntactic and external knowledge which are
crucial for generating responses in a chit-chat
system. To overcome this problem, we propose an end to end multi-stream deep learning
architecture which learns unified embeddings
for query-response pairs by leveraging contextual information from memory networks and
syntactic information by incorporating Graph
Convolution Networks (GCN) over their dependency parse. A stream of this network
also utilizes transfer learning by pre-training
a bidirectional transformer to extract semantic representation for each input sentence and
incorporates external knowledge through the
neighborhood of the entities from a Knowledge Base (KB). We benchmark these embeddings on next sentence prediction task and
significantly improve upon the existing techniques. Furthermore, we use AMUSED to represent query and responses along with its context to develop a retrieval based conversational agent which
has been validated by expert linguists to have
comprehensive engagement with humans.
\end{abstract}

\section{Introduction}

With significant advancements in Automatic speech
recognition systems \citep{hinton2012deep,speech_survey} and the field of natural language
processing, conversational agents have become an important
part of the current research.  It finds its usage in multiple domains ranging from self-driving cars \citep{cars} to social robots and virtual assistants \citep{Chen}. Conversational agents can be broadly classified into two categories:  a task oriented chat bot and a chit-chat based system respectively. The former works towards completion of a certain goal and are specifically designed for domain-specific needs such as restaurant reservations \citep{wen}, movie recommendation \citep{dhingra}, flight ticket booking systems \citep{flight} among many others. The latter is more of a personal companion and engages in human-computer interaction for entertainment or emotional companionship. An ideal chit chat system  should be able to perform non-monotonous interesting conversation with context and coherence.

\begin{figure}[t]
\begin{center}
\includegraphics[width = \textwidth]{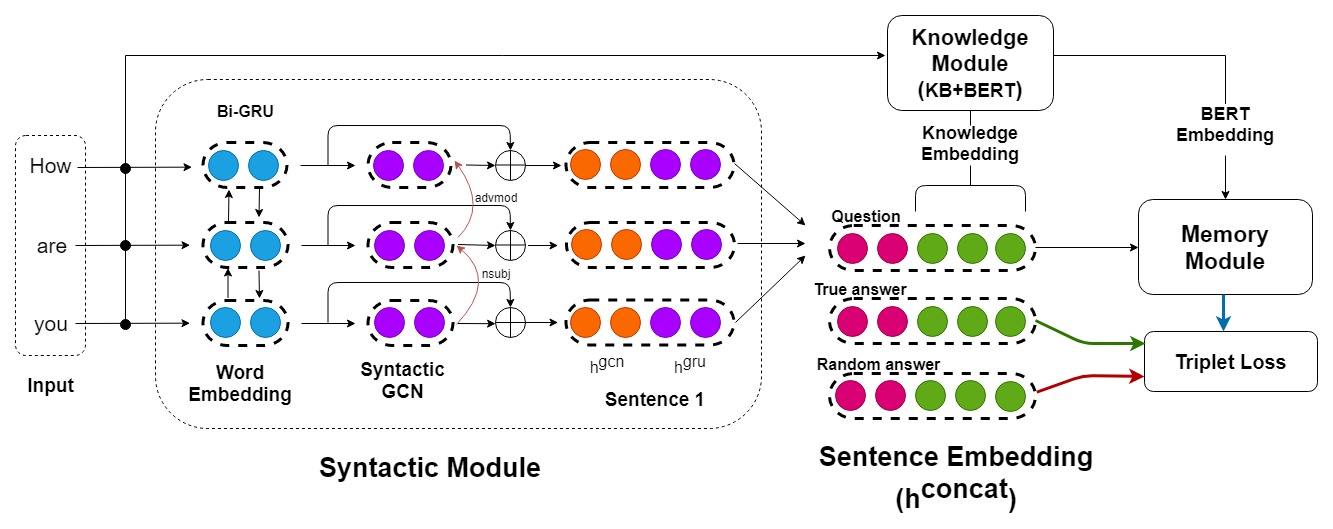}
\end{center}
\caption{\label{fig:model_overview}Overview of \method{}. \method{} first encodes each sentence by concatenating embeddings (denoted by $\oplus$) from Bi-\lstm{} and Syntactic GCN for each token, followed by word attention. The sentence embedding is then concatenated with the knowledge embedding from the Knowledge Module (\reffig{fig:knowledge_module}). The query embedding passes through the Memory Module (\reffig{fig:memory_module}) before being trained using triplet loss. Please see \refsec{sec:details} for more details.}
\end{figure}

Current chit chat systems are either generative \citep{vinyals2015neural} or retrieval based in nature. The generative ones tend to generate natural language sentences as responses and enjoy scalability to multiple domains without much change in the network.
Even though easier to train, they suffer from error-prone responses \citep{zhang2018learning}. IR based methods select the best response from a given set of answers which makes them error-free. But, since the responses come from a specific dataset, they might suffer from distribution bias during the course of conversation.


A chit-chat system should capture semantic, syntactic, contextual and external knowledge in a conversation to model human like performance. Recent work by \citet{bordes2016learning} proposed a
memory network based approach to encode contextual
information for a query while performing generation and retrieval
later. Such networks can capture long term context but fail  to encode relevant syntactic information through their model. Things like anaphora resolution are properly taken care of if we incorporate syntax.

Our work improves upon previous architectures by creating enhanced representations of the conversation using
multiple streams which includes Graph Convolution networks \citep{gcn_first_work}, transformers \citep{transformer} and memory networks \citep{bordes2016learning} in an end to end setting,  where each component captures conversation relevant information from queries, subsequently leading to better responses. Our contribution for this paper can be summarized as follows:

\begin{itemize}
\item We propose \method{}, a novel multi stream deep learning model which learns rich unified embeddings for query response pairs using triplet loss as a training metric.
\item We perform multi-head attention over query-response pairs which has proven to be much more effective than unidirectional or bi-directional attention.
\item  We use Graph Convolutions Networks in a chit-chat setting to incorporate the syntactical information in the dialogue using its dependency parse.
\item Even with the lack of a concrete metric to judge a conversational agent, our embeddings have shown to perform  interesting response retrieval on Persona-Chat dataset.

\end{itemize}

\section{Related Work}
\label{sec:related_work}


The task of building a conversational agent has gained much traction in the last decade with various techniques being tried to generate relevant human-like responses in a chit-chat setting. Previous modular systems \citep{martin2009speech} had a complex pipeline based structure containing various hand-crafted rules and features making them difficult to train. This led to the need of simpler models which could be trained end to end and extended to multiple domains. \citet{vinyals2015neural} proposed a simple sequence to sequence model that could generate answers based on the current question, without needing extensive feature engineering and domain specificity. However, the responses generated by this method lacked context. To alleviate this problem, \citet{Sordoni} introduced a dynamic-context generative network which is shown to have improved performance on unstructured Twitter conversation dataset.  To model complex dependencies between sub-sequences in an utterance, \citet{serban} proposed a hierarchical latent variable encoder-decoder model. It is able to generate longer outputs while maintaining context at the same time. Reinforcement learning based approaches have also been deployed to generate interesting responses  \citep{zhang_rl} and tend to possess unique conversational styles \citep{asghar_rl}.

With the emergence of a number of large datasets, retrieval methods have gained a lot of popularity. Even though the set of responses are limited in this scenario, it doesn't suffer from the problem of generating meaningless responses. A Sequential Matching Network proposed by \citet{wu_sequential} performs word matching of responses with the context before passing their vectors to a RNN. Addition of external information along with the current input sentence and context improves the system as is evident by incorporating a large common sense knowledge base into an end to end conversational agent \citep{young2018augmenting}. To maintain diversity in the responses, \citet{song2018towards} suggests a method to combine a probabilistic model defined on item-sets with a seq2seq model.  Responses like \textit{'I am fine'} can make conversations monotonous; a specificity controlled model \citep{zhang2018learning} in conjunction with seq2seq architecture overcomes this problem. These networks helps solve one or the other problem in isolation.

To maintain proper discourse in the conversation, context vectors are passed together with input query vector into a deep learning model \citep{Sordoni}. A context modelling approach which includes concatenation of dialogue history has also been tried \citep{martin2009speech}. However, the success of memory networks on Question-Answering task \citep{sukhbaatar2015end} opened the door for its further use in conversational agents. \citet{bordes2016learning} used the same in a task oriented setting for restaurant domain and reported accuracies close to $96\%$ in a full dialogue scenario. \citet{persona} further used these networks in a chit chat setting on Persona-Chat dataset and came up with personalized responses.

In our network, we make use of Graph Convolution Networks \citep{kipf2016semi,Defferrard:2016:CNN:3157382.3157527}, which have been found to be quite effective for encoding the syntactic information present in the dependency parse of sentences \citep{gcn_srl}. External Knowledge Bases (KBs) have been exploited in the past to improve the performances in various tasks \citep{cesi_paper,reside2018,figer_paper}. The relation based strategy followed by \citet{hixon} creates a KB from dialogue itself, which is later used to improve Question-Answering \citep{saha}. \citet{postech,kggrounded} have used KBs to generate more informative responses by using properties of entities in the graph.  \citep{young2018augmenting} focused more on introducing knowledge from semantic-nets rather than general KBs. 
\section{Background: Graph Convolution Networks}
\label{sec:gcn_background}
\textbf{GCN for undirected graph: } For an undirected graph $G = (V, E)$, where $V$ is the set of $n$ vertices and $E$ is the set of edges, the representation of the node $v$ is given by
        $x_{v} \in \mathbb{R}^{m}, \forall v \in V$ . The output hidden representation $ h_{v}$ of the node after one layer of GCN is obtained by considering only the immediate neighbors of the node as given by \citet{kipf2016semi}. To capture the multi-hop representation, GCN layers can be stacked on top of each other.

\textbf{GCN for labeled directed graph: } For a directed graph $G = (V, E)$, where $V$ is the set of vertices we define the edge set $E$ as a set of tuples $(u, v, l(u,v))$ where there is an edge having label $l(u,v)$ between nodes $u$ and $v$. \citet{gcn_srl} proposed the assumption that information doesn't necessarily propagate in certain directions in the directed edge, therefore, we add tuples having inverse edges $(v, u, l(u,v)^{-1})$ as well as self loops $(u, u, \Omega)$, where $\Omega$ denotes self loops, to our edge set $E$ to get an updated edge set $E'$. 

The representation of a node $x_{v}$, after the $k^{th}$ layer is given as : \[h_{v}^{k+1} = f \left(\sum_{u \in N(v)} (W_{l(u,v)}^{k}h_{u}^{k} + b_{l(u,v)}^{k}) \right).\] where $W_{l(u,v)}^k \in \mathbb{R}^{d \times d}$ and $b_{l(u,v)}^k \in \mathbb{R}^{d}$ are trainable edge-label specific parameters for the layer $k$, $N(v)$ denotes the set of all vertices that are immediate neighbors of $v$ and $f$ is any non-linear activation function (e.g., ReLU: $f(x) = $ max$(0, x)$). 

Since we are obtaining the dependency graph from Stanford CoreNLP \citep{stanford_corenlp}, some edges can be erroneous. Edgewise gating \citep{gcn_nmt, gcn_srl} helps to alleviate this problem by decreasing the effects of such edges. For this, each edge $(u, v, l(u,v))$ is assigned a score which is given by : \[g^{k}_{uv} = \sigma(h_{u}^{k} \cdot \hat{w}_{l(u,v)}^{k} + \hat{b}_{l(u,v)}^{k}),\] where $\hat{w}_{l(u,v)}^{k} \in \mathbb{R}^{m}$ and $\hat{b}_{l(u,v)}^{k}) \in \mathbb{R}$ are trained and $\sigma$ denotes the sigmoid function. Incorporating this, the final GCN embedding for a node $v$ after $n^{th}$ layer is given as : 
\begin{equation}
\label{eqn:gnc_eqn}
h_{v}^{n+1} = f \bigg(\sum_{u \in N(v)} g^{k}_{uv} \times (W_{l(u,v)}^{n}h_{u}^{n} + b_{l(u,v)}^{n}) \bigg).
\end{equation}
\section{\method{} Details}
\label{sec:details}
This section provides details of three main components of \method{} which can broadly be classified into Syntactic, Knowledge and Memory Module. We hypothesize that each module captures information relevant for learning representations, for a query-response pair in a chit-chat setting.

Suppose that we have a dataset $\mathcal{D}$ consisting of a set of conversations $d_1, d_2, ..., d_C$ where $d_c$ represents a single full length conversation consisting of multiple dialogues. A conversation $d_c$ is given by a set of tuples ${(q_1, r_1), (q_2, r_2), ..., (q_n, r_n)} $ where a tuple $(q_i, r_i)$ denotes the query and response pair for a single turn. The context for a given query $q_i\ \forall\ i \geq 2$ is defined by a list of sentences $l$ : [$q_1, r_1, ..., q_{i-1}, r_{i-1}]$. We need  to find the best response $r_i$ from the set of all responses, $\mathcal{R}$. 
The training set $\mathcal{D}^{'}$ for \method{} is defined by set of triplets $(q_i, r_i, n_i) ~ \forall ~ 1 \leq i \leq N$ where $N$ is the total number of dialogues and $n_i$ is a negative response randomly chosen from set $\mathcal{R}$.

\subsection{Syntactic Module}
\label{sec:syntactic_module_details}

Syntax information from dependency trees has been successfully exploited to improve a lot of Natural Language Processing (NLP) tasks \citep{cesi_paper,mintz2009distant}. In dialog agents, where anaphora resolution as well as sentence structure influences the responses, it finds special usage. A Bi-\gru{} \citep{gru_paper} followed by a syntactic GCN is used in this module.

Each sentence $s$ from the input triplet is represented with a list of $k$-dimensional GloVe embedding \citep{pennington2014glove} corresponding to each of the $m$ tokens in the sentence. The sentence representation $S \in \mathbb{R}^{m \times k}$ is then passed to a Bi-\gru{} to obtain the representation $S^{gru} \in \mathbb{R}^{m \times d_{gru}}$, where $d_{gru}$ is the dimension of the hidden state of Bi-\gru{}. 

This contextual encoding \citep{rnn_speech_recog} captures the local context really well, but fails to capture the long range dependencies that can be obtained from the dependency trees. We use GCN to encode this syntactic information. Stanford CoreNLP \citep{stanford_corenlp} is used to obtain the dependency parse for the sentence $s$. Giving the input as $S^{gru}$, we use GCN  \refeqn{eqn:gnc_eqn}, to obtain the syntactic embedding $S^{gcn}$. Following \citet{gcn_event}, we only use three edge labels, namely forward-edge, backward-edge and self-loop. This is done because incorporating all the edge labels from the dependency graph heavily over-parameterizes the model. 

The final token representation is obtained by concatenating the contextual Bi-\gru{} representation $h^{gru}$ and the syntactic GCN representation $h^{gcn}$. A sentence representation is then obtained by passing the tokens through a layer of word attention \citep{bahdanau+al-2014-nmt} as used by \citep{reside2018,bgwa_paper}, which is concatenated with the embedding obtained from the Knowledge Module (described in \refsec{sec:knowledge_module_details}) to obtain the final sentence representation $h^{concat}$.
\begin{figure}[t]
\centering
\begin{minipage}{.46\textwidth}
\centering
	\includegraphics[width = \textwidth]{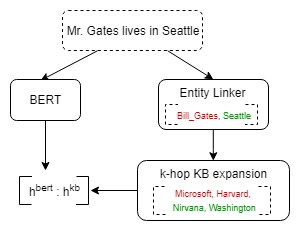}
	\caption{Description of Knowledge Module. The input sentence is passed to a pre-trained BERT model, output from which is concatenated with averaged embedding from the KB-neighbors of entities present in the input. Refer \refsec{sec:knowledge_module_details} for a detailed explanation.}
	\label{fig:knowledge_module}
	\end{minipage}%
	\hfill
\begin{minipage}{.46\textwidth}
\centering
\includegraphics[width=\textwidth]{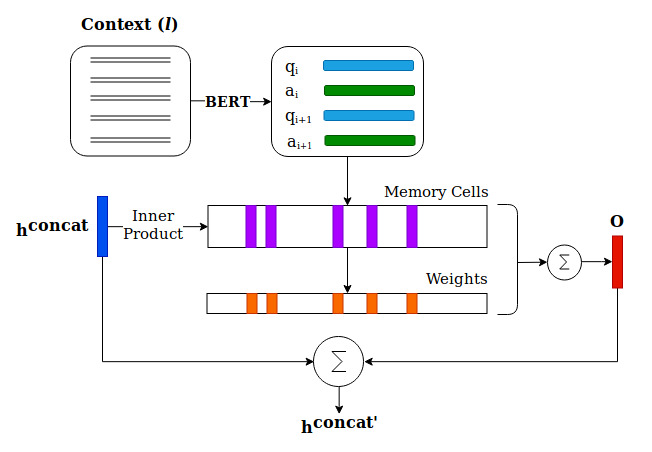}
\caption{\label{fig:memory_module}Memory Module description. The query representation and BERT embeddings of the context sentences is passed to the memory network to capture the dialogue context. Please see \refsec{sec:memory_module_details} for more details.}  
\end{minipage}
\end{figure}

\subsection{Knowledge Module}
\label{sec:knowledge_module_details}
The final sentence representation $h^{concat}$ of the query is then passed into Knowledge Module. It is further subdivided into two components: a pre-trained Transformer model for next dialogue prediction problem and a component to incorporate information from external Knowledge Bases (KBs).

\subsubsection{Next Dialogue Prediction Using Transformers}
\label{bertmodule}
The next dialogue prediction task is described as follows: For each 
query-response pair in the dataset, we generate a positive sample $(q, r)$ and a negative sample $(q, n)$ where $n$ is randomly chosen from the set of responses $\mathcal{R}$ in dataset $D$. Following \citet{bert}, a training example is defined by concatenating $q$ and $r$ which are separated by a delimiter $||$ and is given by $[q||r]$. 
The problem is to classify if the next sentence is a correct response or not.

A pre-trained BERT model is used to further train a binary classifier for the next dialogue prediction task as described above. After the model is trained, the pre-final layer is considered and the vector from the special $cls$ token is chosen as the sentence representation. The representation thus obtained would have a tendency to be more inclined towards its correct positive responses. Multi-head attention in the transformer network, along with positional embeddings during training, helps it to learn intra as well as inter sentence dependencies \citep{bert,transformer}. The input query sentence is then passed from this network to obtain the BERT embedding, $h^{bert}$.

\subsubsection{k-Hop KB Expansion}
In our day-to-day conversations, to ask succinct questions, or to keep the conversation flowing, we make use of some background knowledge
. For example, if someone remarks that they like rock music, we can ask a question if they have listened to Nirvana. 
It can be done only if we know that Nirvana plays rock music. To incorporate such external information, we can make use of existing Knowledge Bases like Wikipedia, Freebase \citep{freebase_paper} and Wikidata \citep{wikidata_paper}. Entities in these KBs are linked to each other using relations. We can expand the information we have about an entity by looking at its linked entities. Multiple hops of the Knowledge Graph (KG) can be used to expand knowledge.

In \method{}, we do this by passing the input query into Stanford CoreNLP to obtain entity linking information to Wikipedia. Suppose the Wikipedia page of an entity $e$ contains links to the set of entities $E$. We ignore relation information and only consider one-hop direct neighbors of $e$. To obtain a KB-expanded embedding $h^{kb}$ of the input sentence, we take the average of GloVE embeddings of each entity in $E$. 
In place of Wikipedia, bigger knowledge bases like Wikidata, as well as relation information, can be used to improve KB embeddings. We leave that for future work.

\subsection{Memory Module}
\label{sec:memory_module_details}
For effective conversations, it is imperative that we form a sense from the dialogues that have already happened. A question about '\textit{Who is the president of USA}' followed by '\textit{What about France}' should be self-containing. This dialogue context is encoded using a memory network \citep{sukhbaatar2015end}. The memory network helps to capture context of the conversation by storing dialogue history i.e. both question and responses. 
The query representation, $h^{concat}$ is passed to the memory network, along with BERT embeddings $h^{bert}$ of the context, from the Knowledge Module (\refsec{sec:knowledge_module_details}). 

In \method{}, memory network uses supporting memories to generate the final query representation $(h^{concat'})$. Supporting memories contains input $(m_{i})$ and output $(c_{i})$ memory cells \citep{sukhbaatar2015end}. The incoming query $q_i$ as well as the history of dialogue context $l: [(q_{1},r_{1}),..,(q_{i-1},r_{i-1})]$
is fed as input. The memory cells are populated using the BERT representations of context sentences $l$ as follows:
$ m_{i} = c_{i} = \{BERT(x),\, x \in \tau\}$, where $\tau = [q_1, r_1, q_2, r_2, ..., q_i, r_i]\  \forall\  (q_i, r_i) \in l$.

Following \citet{bordes2016learning}, the incoming query embedding along with input memories is used to compute relevance of context stories as a normalized vector of attention weights as $ a_{i} = \norm{( < m_{i} , h^{concat}> )}$, where $<a,b>$ represents the inner product of $a$ and $b$. The response from the output memory, $o$, is then generated as : $ o = \sum_{i} a_{i}c_{i}$. The final output of the memory cell, $u$ is obtained by adding $o$ to $h^{concat}$. To capture context in an iterative manner, memory cells are stacked in layers \citep{sukhbaatar2015end} which are called as hops. The output of the memory cell after the $k^{th}$ hop is given by $u^{k} = o^{k-1} + u^{k-1}$ where $u^{0} = h^{concat}$.
The memory network performs $k$ such hops and the final representation $h^{concat'}$ is given by sum of $o^k$ and $u^{k}$.
%
\subsection{Triplet Loss}
Triplet loss has been successfully used for face recognition \citep{facenet}. Our insight is that traditional loss metrics might not be best suited for a retrieval-based task with a multitude of valid responses to choose from. We define a \textit{Conversational Euclidean Space} where the representation of a sentence is driven by its context in the dialogue along with its syntactic and semantic information.  We have used this loss to bring the query and response representations closer in the conversational space. Questions with similar answers should be closer to each other and the correct response.
An individual data point is a triplet which consists of a query $(q_i)$, its correct response $(r_i)$ and a 
negative response $(n_i)$ selected randomly. We need to learn  their embeddings $\phi(q_i) = h^{concat'}_{q_i}$, $\phi(r_i) = h^{concat}_{r_i}$ and $\phi(n_i) = h^{concat}_{n_i}$ such that the positive pairs are closer in the embedding space compared to the negative ones. This leads to the following equation:
\[ \norm { \phi (q_i) -\phi (r_i)}_2^2 + \alpha < \norm { \phi (q_i) -\phi (n_i)}_2^2 ,\]
where $\alpha$ is the margin hyper-parameter used to separate negative and positive pairs.
If $I$ be the set of triplets, $N$ the number of triplets and $w$ the parameter set, then, triplet loss $(\mathcal{L})$ is defined as :  \[\mathcal{L}(I,w) = \sum_{i=0}^{N} [ \norm { \phi (q_i) -\phi (r_i)}_2^2 -\norm {\phi (q_i) -\phi (n_i)}_2^2 + \alpha]_{+}  \]

\section{Experimental Setup}

\subsection{Datasets}
\label{sec:datasets}
\textbf{Persona-Chat}: We use this dataset to build and evaluate the chit-chat system. Persona-Chat \citep{persona} is an open domain dataset on personal conversations created by randomly pairing two humans on Amazon Mechanical Turk.
The paired crowd workers converse in a natural manner for $6-12$ turns. This made sure that the data mimic normal conversations between humans which is very crucial for building such a system. This data is not limited to social media comments or movie dialogues. It contains $9907$ training conversations and $1000$ conversations each for testing and validation. There are a total of $131,438$ query-response pairs with a vocab size of $19262$ in the dataset. We use  it for training \method{} as it provides consistent conversations with proper context. 

\textbf{DSTC}: Dialogue State Tracking Challenge dataset \citep{henderson} contains conversations for restaurant booking task. Due to its task oriented nature, it doesn't need an external knowledge module, so we train it only using memory and syntactic module and test on an automated metric.

\textbf{MNLI} and \textbf{MRPC}: We further use Multi-Genre Natural Language Inference and Microsoft Research Paraphrase Corpus \citep{wang2019glue} to fine-tune parts of the network i.e; Knowledge Module. It is done because these datasets resemble the core nature of our problem where in we want to predict the correctness of one sentence in response to a particular query.

\begin{table}[t]
\begin{minipage}{0.56\textwidth}
\caption{\label{tb:prevdatasets} Accuracy as an automatic evaluation metric on Next Dialogue Prediction task over Persona Chat. We perform different operations on embeddings of sentence pairs to study ablation. Concat, Diff and Min refers to Concatenation, Difference and Element wise min respectively. See \refsec{subsec:ablation} for more details.}

	\label{Table2}
    \begin{center}
    \begin{adjustbox}{width=\textwidth}
	\begin{tabular}{ccccc}
		\toprule
		\textbf{Model}     & \textbf{Concat} & \textbf{Diff }  & \textbf{Min}  \\
		\midrule
		Bi-GRU \& GCN only & 72.8\% & 73.2\% & 59.4\%  \\
		
		\midrule
		BERT only & 74.3\% & 71.9\% & 66.3\%  \\
		
		\midrule
		BERT with Bi-GRU \\ \& GCN & 77.7\% & 73.3\% & 71.4\%  \\
		
		\midrule
		BERT and KB with \\Bi-GRU
		and memory networks & \textbf{83.6\%} & 78.4\% & 69.2\% \\

		\bottomrule
	\end{tabular}
	\end{adjustbox}
	\end{center}
	\end{minipage}
\hspace{1cm}
\begin{minipage}{0.34\textwidth}	
\caption{Precision @1 comparison between different methods. Precision@1 \% tell us the number of times the correct response from the dataset comes up. Details in  \refsec{subsec:ablation}}

	\label{Table3}
    \begin{center}
     \begin{adjustbox}{width=\textwidth}
	\begin{tabular}{cc}
		\toprule
		\textbf{Method}     & \textbf{Precision@1}  \\
		\midrule
		Seq2Seq & 0.092  \\
		
		\midrule
		Profile Memory & 0.092  \\
		
		\midrule
		IR Baseline  & 0.214  \\
		
		\midrule
		AMUSED(Persona Chat) & \textbf{0.326} \\
		
		\midrule
		AMUSED(DSTC) & \textbf{0.78} \\
		\bottomrule
		\addlinespace
	\end{tabular}
	\end{adjustbox}
	\end{center}
	\end{minipage}
\end{table}

\subsection{Training}
\label{sec:baselines}

\textbf{Pre-training BERT}:
Before training \method{}, the knowledge module is processed by pre-training a bidirectional transformer network and extracting one hop neighborhood entities from Wikipedia KB. 
We use the approach for training as explained in \refsec{bertmodule}. There are 104224 positive training and 27214 validation query-response pairs from Persona Chat. We perform three different operations: a) Equal sampling: Sample equal number of negative examples from dataset, b) Oversampling: Sample double the negatives to make training set biased towards negatives and c) Under sampling: Sample $70\%$ of negatives to make training set biased towards positives. Batch size and maximum sequence length are 32 and 128 respectively.
We fine-tune this next sentence prediction model with MRPC and MNLI datasets which improves the performance.


\textbf{Training to learn Embeddings}: 
\method{} requires triplets to be trained using triplet loss. A total of $131438$ triplets of the form $(q,r,n)$ are randomly split in 90:10 ratio to form training and validation set. The network is trained with a batch size of $64$ and dropout of 0.5. Word embedding size is chosen to be 50. Bi-GRU and GCN hidden state dimensions are chosen to be 192 and 32 respectively. One layer of GCN is employed. Validation loss is used as a metric to stop training which converges after $50$ epochs using Adam optimizer at 0.001 learning rate.

\subsection{Retrieval}
As a retrieval based model, the system selects a response from the predefined answer set. The retrieval unit extracts embedding $(h^{concat})$ for each answer sentence from the trained model and stores it in a representation matrix which will be utilized later during inference.

First, a candidate subset $A$ is created by sub-sampling a set of responses having overlapping words with a given user query. Then, the final output is retrieved on the basis of cosine similarity between query embedding $h^{concat'}$ and the extracted set of potential responses $(A)$. 
The response with the highest score is then labelled as the final answer and the response embedding is further added into the memory to take care of context.

\subsection{Results And Evaluation}
\subsubsection{Selecting The Pre-Trained Model}
The model resulting from oversampling method  beats its counterparts by \textbf{more than $\bf 3\%$ }in accuracy. It clearly indicates that a better model is one which learns to distinguish negative examples well. The sentence embeddings obtained through this model is further used for lookup in the Knowledge Module (\refsec{sec:knowledge_module_details}) in \method{}.

\subsubsection{Ablation Studies on Automated Metrics}
\label{subsec:ablation}

We use two different automated metrics to check the effectiveness of the model and the query-response representations that we learnt.

\textbf{Next Dialogue Prediction Task}: Various components of \method{} are analysed for their performance on next dialogue prediction task. This task tell us that, given two sentences (a query and a response) and the context, whether second sentence is a valid response to the first sentence or not. Embeddings for queries and responses are extracted from our trained network and then multiple operations which include a) Concatenation, b) Element wise min and c) Subtraction are performed on those before passing them to a binary classifier. A training example consists of embeddings of two sentences from a $(q,a)$ or $(q,n)$ pair which are created in a similar fashion as in \refsec{bertmodule}.

Accuracy on this binary classification problem has been used to select the best network. Furthermore, we perform ablation studies using different modules to understand the effect of each component in the network. A 4 layer neural network with ReLU activation  in its hidden layers and softmax in the final layer is used as the classifier. External knowledge in conjunction with memory and GCN module has the best accuracy when embeddings of query and response are concatenated together. A detailed study of performance of various components over these operations is shown in Table \ref{tb:prevdatasets}.

\textbf{Precision@1}: This is another metric used to judge the effectiveness of our network. It is different from the next sentence prediction task accuracy. It measures that for n trials, the number of times a relevant response is reported with the highest confidence value. \reftable{Table2} reports a comparative study of this metric on 500 trials conducted for AMUSED along with results for other methods. DSTC dataset is also evaluated on this metric without the knowledge module as explained in \refsec{sec:datasets}

Looking for exact answers might not be a great metric as many diverse answers might be valid for a particular question. So, we must look for answers which are contextually relevant for that query. Overall, we use next sentence prediction task accuracy to choose the final model before retrieval.
\begin{table}[t]
\caption{\label{tb:datasets}Human based evaluation is conducted for 5 different components in the network as well as KV memory networks. AMUSED achieves the highest percent gain over specified baseline model. The scale is 1-10.}
\label{sample-table}
\begin{center}
\begin{adjustbox}{width=\textwidth}
\begin{tabular}{cccccc}
\toprule
		\textbf{Model}     & \textbf{Coherence} & \textbf{Context Aware}  & \textbf{Non Monotonicity }& \textbf{Average Rating} & \textbf{\%gain}\\
		
		\midrule
		Bi-GRU \& GCN only & 6.82 & 7.35 & 6.77 & 6.98 & Baseline  \\
		
		\midrule
		BERT only & 7.61 & 7.24 & 6.33 & 7.06 & 1.14  \\
		
		\midrule
		BERT with Bi-GRU \& GCN & 7.54 & 6.91 & 7.38 & 7.27 & 4.15  \\
		\midrule
		BERT and External KB \\with Bi-GRU \& GCN & 7.16 & 7.34 & 7.72 & 7.40 & 6.01  \\
		\midrule
		KV Memory Networks\citep{persona} & 7.56 & 8.09 & 7.84 & 7.83 & 12.18  \\
		
		\midrule
		BERT \& External KB with Bi-GRU,\\ GCN
		and memory networks & 8.21 & 8.34 & 7.82 & 8.12 & 16.33  \\

		\bottomrule
\end{tabular}
\end{adjustbox}
\end{center}
\end{table}

\subsubsection{Ablation Study by Humans}

There is no concrete metric to evaluate the performance of an entire conversation in a chit-chat system. Hence, human evaluation was conducted using expert linguists to check the quality of conversation. They were asked to chat for $7$ turns and rate the quality of responses on a scale of $1-10$ 
where a higher score is better. Similar to \citet{persona}, there were multiple parameters to rate the chat based on coherence, context awareness and non-monotonicity to measure various factors that are essential for a natural dialogue. By virtue of our network being retrieval based, we don't need to judge the responses based on their structural correctness as this will be implicit.

To monitor the effect of each neural component, we get it rated by experts either in isolation or in conjunction with other components. Such a study helps us understand the impact of different modules on a human based conversation. Dialogue system proposed by \citet{persona} is also reproduced and reevaluated for comparison.
 From Table \ref{tb:datasets} we can see that human evaluation follows a similar trend as the automated metric, with the best rating given to the combined architecture. 
\section{Conclusion}
\label{sec:conclusion}
In the paper, we propose \method{}, a multi-stream architecture which effectively encodes semantic information from the query while properly utilizing external knowledge for improving performance on natural dialogue. It also employs GCN to capture long-range syntactic information and improves context-awareness in dialogue by incorporating memory network. Through our experiments and results using different metrics, we demonstrate that learning these rich representations through smart training (using triplets) would improve the performance of chit-chat systems. The ablation studies show the importance of different components for a better dialogue. Our ideas can easily be extended to various conversational tasks which would benefit from such enhanced representations.

\bibliography{iclr2020_conference}
\bibliographystyle{iclr2020_conference}


\end{document}